\pdfoutput=1

\documentclass[11pt,table]{article}

\usepackage[final]{templatepackage/acl}

\usepackage{times}
\usepackage{latexsym}

\usepackage[T1]{fontenc}

\usepackage[utf8]{inputenc}

\usepackage{microtype}

\usepackage{inconsolata}

\usepackage{graphicx}

\usepackage{physics}
\usepackage{amsfonts}
\ifcsname proof\endcsname
\else
\usepackage{amsthm}
\fi
\usepackage{amssymb}
\usepackage{bm}
\usepackage{cleveref}
\usepackage{hhline}
\usepackage{multirow}
\usepackage{enumitem}
\ifcsname algorithm\endcsname\else
\usepackage{algorithm}
\usepackage{algorithmic}
\fi
\ifcsname cellcolor\endcsname\else\usepackage[table]{xcolor}\fi
\ifcsname mathclap\endcsname\else\usepackage{mathtools}\fi
\usepackage{booktabs}
\usepackage{makecell} \newcommand{\opn}[1]{\operatorname{#1}}
\allowdisplaybreaks
\makeatletter
\newcommand{\myvspace}{\@ifstar\myvspacestar\myvspacenostar}
\newcommand{\myvspacenostar}[1]{}
\newcommand{\myvspacestar}[1]{}
\makeatother \ifcsname theorem\endcsname
\else
    
\fi
\ifcsname definition\endcsname
\else
    
\fi
\ifcsname lemma\endcsname
\else
    
\fi %
  \usepackage{tikz}
\usetikzlibrary{shapes, arrows.meta, positioning}
\usepackage{fontawesome5}
\usepackage{pifont}
\usepackage{pgfplots}  
\title{A Bayesian Approach to Harnessing the Power of LLMs in Authorship Attribution}

\author{
 \textbf{Zhengmian Hu\textsuperscript{1,2$\ast$}},
 \textbf{Tong Zheng\textsuperscript{1$\ast$}},
 \textbf{Heng Huang\textsuperscript{1}}
\\
\\
 \textsuperscript{1}Department of Computer Science, University of Maryland, College Park, MD 20742\\
 \textsuperscript{2}Adobe Research
\\
\small{
   \href{mailto:huzhengmian@gmail.com}{huzhengmian@gmail.com},
   \href{mailto:zhengtong12356@gmail.com}{zhengtong12356@gmail.com},
   \href{mailto:heng@umd.edu}{heng@umd.edu},
 }
}

\begin{document}
\maketitle
\def\thefootnote{*}\footnotetext{These authors contributed equally to this work.}
\begin{abstract}
Authorship attribution aims to identify the origin or author of a document. Traditional approaches have heavily relied on manual features and fail to capture long-range correlations, limiting their effectiveness. Recent advancements leverage text embeddings from pre-trained language models, which require significant fine-tuning on labeled data, posing challenges in data dependency and limited interpretability. Large Language Models (LLMs), with their deep reasoning capabilities and ability to maintain long-range textual associations, offer a promising alternative. This study explores the potential of pre-trained LLMs in one-shot authorship attribution, specifically utilizing Bayesian approaches and probability outputs of LLMs. Our methodology calculates the probability that a text entails previous writings of an author, reflecting a more nuanced understanding of authorship. By utilizing only pre-trained models such as Llama-3-70B, our results on the IMDb and blog datasets show an impressive 85\% accuracy in one-shot authorship classification across ten authors. Our findings set new baselines for one-shot authorship analysis using LLMs and expand the application scope of these models in forensic linguistics. This work also includes extensive ablation studies to validate our approach.
\end{abstract} \section{Introduction}

Authorship attribution, the process of identifying the origin or author of a document, has been a longstanding challenge in forensic linguistics. It has numerous applications, including detecting plagiarism \cite{alzahrani2011understanding} and attribution of historical text \cite{silva2023authorship}. As the digital age progresses, the need for reliable methods to determine authorship has become increasingly important, especially in the context of combating misinformation spread through social media and conducting forensic analysis. The ability to attribute authorship can also lead to challenges around privacy and anonymity \cite{juola2008authorship}.

The field traces its roots back to the early 19th century \cite{mechti2021orderly}, with early studies focusing on stylistic features and human expert analysis \cite{mosteller1963inference}. Traditional methods often relied on stylometry, which quantifies writing styles \cite{holmes1994authorship}, and rule-based computational linguistic methods \cite{stamatatos2009survey} to deduce authorship. Later, statistical algorithms incorporating extensive text preprocessing and feature engineering \cite{bozkurt2007authorship,seroussi2014authorship} were introduced to improve accuracy. However, these methods often struggled with capturing long-range dependencies in text and require careful setup of specific thresholds for various indicators, which can be challenging to select effectively. They also involve designing complex, high-quality features, which can be costly and time-consuming.

The advent of deep learning has transformed the landscape of authorship attribution by turning the problem into a multi-class classification challenge, allowing for the capture of more features and addressing more complex scenarios effectively~\cite{ruder2016character,ge2016authorship, shrestha2017convolutional, zhang-etal-2018-syntax}. However, these neural network (NN) models often lack interpretability and struggle with generalization in cases of limited samples.

Despite advancements, the field still faces significant challenges. Obtaining large, balanced datasets that represent multiple authors fairly is difficult, and as the number of authors increases, the accuracy of machine learning models tends to decrease.

On the other hand, language models, central to modern NLP applications, define the probability of distributions of words or sequences of words and have traditionally been used to predict and generate plausible language. Yet, for a long time, these models, including high-bias models like bag-of-words and n-gram models, struggled to fit the true probability distributions of natural language. Deep learning's rapid development has enabled orders of magnitude scaling up of computing and data, facilitating the use of more complex models such as Random Forests \cite{breiman2001random}, character-level CNNs \cite{zafar2020language}, Recurrent Neural Networks \cite{bagnall2015author}, and Transformer \cite{vaswani2017attention}.

The recent rapid evolution of Large Language Models (LLMs) has dramatically improved the ability to fit natural language distributions. Trained on massive corpora exceeding 1 trillion tokens, these models have become highly capable of handling a wide range of linguistic tasks, including understanding, generation, and meaningful dialogue \cite{liang2022holistic, bubeck2023sparks, zhang2023prompting, zhang2024benchmarking}. They can also explain complex concepts and capture subtle nuances of language. They have been extensively applied in various applications such as chatbots, writing assistants, information retrieval, and translation services. More impressively, LLMs have expanded their utility to novel tasks without additional training, simply through the use of prompts and in-context learning \cite{brown2020language}. This unique ability motivates researchers to adapt LLMs to an even broader range of tasks and topics including reasoning \cite{wei2022chain}, theory of mind \cite{kosinski2023theory} and medical scenario \cite{singhal2023large}.

Interestingly, language models have also been explored for authorship attribution \cite{agun2017document,le2014distributed,mccallum1999multi}. Recently, research has utilized LLMs for question answering (QA) tasks within the application of authorship verification and authorship attribution \cite{huang2024can}, though these have primarily been tested in small-scale settings. Other approaches have attempted to leverage model embeddings and fine-tuning for authorship attribution, such as using GAN-BERT \cite{silva2023authorship} and BERTAA \cite{fabien-etal-2020-bertaa}. However, these techniques often face challenges with scalability and need retraining when updating candidate authors. Moreover, they require relatively large dataset and multiple epochs of fine-tuning to converge. Given the challenges with current approaches, a natural question arises: \textit{How can we harness LLMs for more effective authorship attribution?}

Two aspects of evidence provide insights to answer the above questions. First, recent studies on LLMs have shown that these models possess hallucination problems \cite{ji2023survey}. More interestingly, the outputs of LLMs given prompts may disagree with their internal thinking \cite{liu-etal-2023-cognitive}. Therefore, it is advisable not to rely solely on direct sampling result from LLMs. Second, the training objective of LLMs is to maximize the likelihood of the next token given all previous tokens. This indicates that probability may be a potential indicator for attributing texts to authors.

Language models are essentially probabilistic models, but we find the probabilistic nature of LLMs and their potential for authorship identification remains underexploited. Our study seeks to bridge this gap. Specifically, we explore the capability of LLMs to perform one-shot authorship attribution among multiple candidates. 

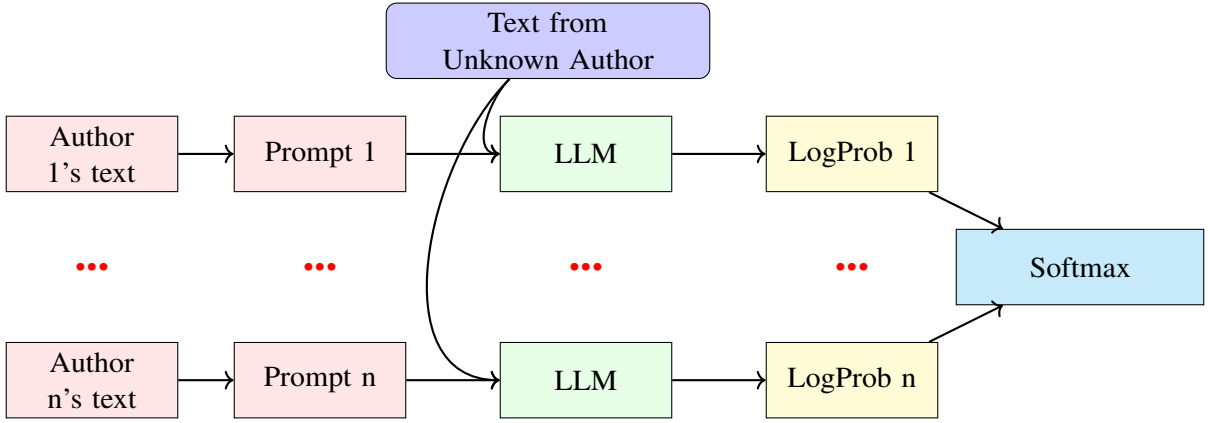
\begin{figure*}
    \centering
\begin{tikzpicture}
\node (unknown) [draw, rectangle, fill=blue!20, text width=4cm, text centered, rounded corners, minimum height=1cm] at (3,3.5) {Text from \\ Unknown Author};

        \node (prompt1) [draw, rectangle, fill=red!10, text centered, minimum height=1cm, text width=2cm] at (0,2) {Prompt 1};
        \node (llm1) [draw, rectangle, fill=green!10, text centered, minimum height=1cm, text width=2cm] at (3.5,2) {LLM};
        \node (prob1) [draw, rectangle, fill=yellow!20, text centered, minimum height=1cm, text width=2cm] at (7,2) {LogProb 1};
        \node (text1) [draw, rectangle, fill=red!10, text centered, minimum height=1cm, text width=2cm] at (-3,2) {Author 1's text};

\node (ellipsis) [text centered, font=\LARGE\bfseries, color=red] at (0,0.5) {...};  \node (ellipsis) [text centered, font=\LARGE\bfseries, color=red] at (-3,0.5) {...};
        \node (ellipsis) [text centered, font=\LARGE\bfseries, color=red] at (3.5,0.5) {...};
        \node (ellipsis) [text centered, font=\LARGE\bfseries, color=red] at (7,0.5) {...};

        \node (prompt3) [draw, rectangle, fill=red!10, text centered, minimum height=1cm, text width=2cm] at (0,-1) {Prompt n};
        \node (llm3) [draw, rectangle, fill=green!10, text centered, minimum height=1cm, text width=2cm] at (3.5,-1) {LLM};
        \node (prob3) [draw, rectangle, fill=yellow!20, text centered, minimum height=1cm, text width=2cm] at (7,-1) {LogProb n};
        \node (text3) [draw, rectangle, fill=red!10, text centered, minimum height=1cm, text width=2cm] at (-3,-1) {Author n's text};

\node (class) [draw, rectangle, fill=cyan!20, text centered, minimum height=1cm, text width=3cm] at (10,0.5) {Softmax};

\draw[->, thick] (text1) -- (prompt1);
        \draw[->, thick] (prompt1) -- (llm1);
        \draw[->, thick] (llm1) -- (prob1);
        \draw[->, thick] (text3) -- (prompt3);
        \draw[->, thick] (prompt3) -- (llm3);
        \draw[->, thick] (llm3) -- (prob3);

\draw[->, thick] (unknown) to[out=225, in=180] (llm1);
        \draw[->, thick] (unknown) to[out=225, in=180] (llm3);

\draw[->, thick] (prob1) -- (class);
        \draw[->, thick] (prob3) -- (class);
    \end{tikzpicture}
\caption{Illustration of bayesian authorship attribution using LLM.}
        \label{fig:illustration}
\end{figure*}

We propose a novel approach based on a Bayesian framework that utilizes the probability outputs from LLMs. By deriving text-level log probabilities from token-level log probabilities, we establish a reliable measure of likelihood that a query text was written by a specific author given example texts from each candidate author. We also design suitable prompts to enhance the accuracy of these log probabilities. By calculating the posterior probability of authorship, we can infer the most likely author of a document (\Cref{fig:illustration}). Due to the pivotal role of log probability in our algorithm, we coined our approach the "Logprob method."

Our new method has three main advantages:

\begin{itemize}[leftmargin=0.15in]
\item No Need for Fine-Tuning: Our approach aligns the classification task with the pretraining objective, both focusing on computing entailment probability. This avoids any objective mismatch introduced by fine-tuning. Moreover, our method leverages the inherent capabilities of pre-trained LLMs and avoids knowledge forgetting that often occurs during fine-tuning.

\item Speed and Efficiency: This approach requires only a single forward pass through the model for each author, making it significantly faster and more cost-effective compared to normal question-answering method of language models which involves sampling a sequence of tokens as answer, with one forward pass for each token generated.

\item No Need for Manual Feature Engineering: The pre-training on diverse data enables LLMs to automatically capture and utilize subtle nuances in language,  thus eliminating the need for manually designing complex features, which can be costly and time-consuming.
\end{itemize}

By applying this technique, we have achieved state-of-the-art results in one-shot learning on the IMDb and blog datasets, demonstrating an impressive 85\% accuracy across ten authors. This advancement establishes a new baseline for one-shot authorship analysis and illustrates the robust potential of LLMs in forensic linguistics. \section{Method}
\newcommand{\localauthortextone}{Tina Fey is a successful professional who has missed out on the baby wagon . All her friends have families and she has promotions . Desperate for a child she tries a sperm bank but it fails when she is told that she is infertile . In desperation she takes on a surrogate who turns her life upside down . Clearly Tina Fey is the smartest one in the room and she walks through this film seemingly on autopilot
and above to everyone around her . What is she doing here ? She is somewhere beyond this film and it shows . Its cute and amusing but Fey's demeanor promises something on a different plane then the rest of the movie . I think the best way to explain it , or over explain it would be Cary Grant in a Three Stooges movie . I think Fey can do great things if she wants or can find material that matches her abilities . A good little film .}
\newcommand{\localauthortexttwo}{In the run-up to the 1972 elections , Washington Post reporter Bob Woodward covers what seems to be a minor break-in at the Democratic Party National headquarters . He is surprised to find top lawyers already on the defence case , and the discovery of names and addresses of Republican fund organisers on the accused further arouses his suspicions . The editor of the Post is prepared to run with the story and assigns Woodward and Carl Bernstein to it . They find the trail leading higher and higher in the Republican Party , and eventually into the White House itself . . . . whatever peoples opinions on the Watergate ' scandel ' , whether they believe it was a big cover up , or the media got a lot wrong , no one can deny just how powerful and interesting this film really is . Pakula directs this very slickly and brings the tension on the two main protagonists very slowly throughout the duration of the movie . Redford and Hoffman work really well together and are given great support from the rest of the cast . the narration works amazingly well and there is good use of mise en scene and connotations . for example there are a few scenes with the t . v screen in the foreground showing Nixon winning his presidential seat again , with Redford in the background , working on his story , probably referring that Nixon always had the scandal at the back of his mind . Pakula makes good use of light and camera-work whenever we meet the elusive ' deep throat ' , and makes the paranoid tension very high with the images of empty streets and a silent soundtrack . very well produced , well acted and a beautiful screenplay , this is a wondrous movie}
\newcommand{\localtesttext}{Barbet Schroeder's portrait of French attorney Jacques Vergès . You've seen him defending people like Klaus Barbie , Carlos the Jackal , Pol Pot as well as other dictators and terrorists . This is a complex story of a complex man and it essentially tells the tale of the man from World War 2 until today . ( And even at 140 minutes the film leaves a great deal out ) . Here is man of his time , who met and defended with many of the famous and infamous people of the last fifty years . He seems to be a man who generally believes in the right of the oppressed to stand up to their oppressors and to have some one to stand up for them . However this is not just the story of a man who fights for the oppressed but it is also the story of a man entangled in things that will cause many to question just how slick a guy is Verges . Many of the terrorists and dictators he defends are in fact his friends , and he is not doing it for the love of cause but also for the love of the finer things . I liked the film a great deal . To be certain I was lost as to bits of the history and who some people were , but at the same time the film isn't about the history , so much as Verges moving through it . This is the story of the man , his causes and to some degree his women . What exactly are we to make of Verges ? I don't know , but I sure do think that he and his life make for a compelling tale . I loved that my idea of what Verges is changed . I loved that I was completely confused at the end as to what I thought , confused in a way that only a film that forces you to think can do . In the end I don't know what I think of Verges, and I love that I will have to sit and reflect on what transpired on screen and in the man's life for a good long while . Certainly one of the better feature length theatrical documentaries to come down the pike in a while . See it . . . probably more than once . See it and then discuss it , it will get the gray cells of your brain working . .}

\newcommand{\localauthortextoneshort}{Tina Fey is a successful professional who has missed out on the baby wagon . All her friends have families and she has promotions . Desperate for a child she tries a sperm bank but it fails when she is told that she is infertile . In desperation she takes on a surrogate who turns her life upside down . Clearly Tina Fey is the smartest one in the room and she walks through this film seemingly on autopilot
and above to everyone around her . What is she doing here ? She is somewhere beyond this film and it shows . Its cute and amusing but Fey's demeanor promises something on a different plane then the rest of the movie . I think the best way to explain it , or over explain it would be Cary Grant in a Three Stooges movie . I think Fey can do great things if she wants or can find material that matches her abilities . A good little film .}
\newcommand{\localauthortexttwoshort}{In the run-up to the 1972 elections , Washington Post reporter Bob Woodward covers what seems to be a minor break-in at the Democratic Party National headquarters . He is surprised to find top lawyers already on the defence case , and the discovery of names and addresses of Republican fund organisers on the accused further arouses his suspicions . The editor of the Post is prepared to run with the story and assigns Woodward and Carl Bernstein to it . They find the trail leading higher and higher in the Republican Party , and eventually into the White House itself . . . . whatever peoples opinions on the Watergate ' scandel ' , whether they believe it was a big cover up , or the media got a lot wrong , no one can deny just how powerful and interesting this film really is . Pakula directs this very slickly and brings the tension on the two main protagonists very slowly throughout the duration of the movie . Redford and Hoffman work really well together and are given great support from the rest of the cast . the narration works amazingly well and there is good use of mise en scene and connotations . for example there are a few scenes with the t . v screen in the foreground showing Nixon winning his presidential seat again , with  \dots}
\newcommand{\localtesttextshort}{Barbet Schroeder's portrait of French attorney Jacques Vergès . You've seen him defending people like Klaus Barbie , Carlos the Jackal , Pol Pot as well as other dictators and terrorists . This is a complex story of a complex man and it essentially tells the tale of the man from World War 2 until today . ( And even at 140 minutes the film leaves a great deal out ) . Here is man of his time , who met and defended with many of the famous and infamous people of the last fifty years . He seems to be a man who generally believes in the right of the oppressed to stand up to their oppressors and to have some one to stand up for them . However this is not just the story of a man who fights for the oppressed but it is also the story of a man entangled in things that will cause many to question just how slick a guy is Verges . Many of the terrorists and dictators he defends are in fact his friends , and he is not doing it for the love of cause but also for the love of the finer things . I liked the film a great deal . To be certain I was lost as to bits of the history and who some people were , but at the same time the film isn't about the history , so much as Verges moving through it . This is the story of the man , his causes and to some degree his women . What exactly are we to make of Verges ? I don't know , but I sure do think that he and his life make for a compelling tale . I loved that my idea of what Verges is changed . I loved that I was completely confused at the end as to what I thought , confused in a way that only a film that forces you to think can do . In the end I don't know what I think  \dots}

\begin{figure*}
\vspace{-15pt}
\centering
\begin{tikzpicture}
\tikzset{
    node distance=0.1cm and 0.1cm,
}

\node[text width=0.45\textwidth, align=center] (author1) {Author 1:};
\node[text width=0.45\textwidth, below=of author1, align=left,font=\tiny] (input1) {
\localauthortextoneshort\\~\\
Here is the text from the same author:\\~\\~\\
{\color{orange}\localtesttextshort}};

\node[text width=0.45\textwidth, right=of author1, align=center] (author2) {Author 2:};
\node[text width=0.45\textwidth, below=of author2, align=left,font=\tiny] (input2) {
\localauthortexttwoshort\\~\\
Here is the text from the same author:\\~\\~\\
{\color{orange}\localtesttextshort}};

\node[below=of input2, align=center] (score2) {-964.51};
\node[align=center] (score1) at (input1 |- 52,52 |- score2) {-958.41};

\node[below=of score1, align=center, text=green] (mark1) {\ding{51}}; \node[below=of score2, align=center, text=red] (mark2) {\ding{55}}; 

\node[left=of mark1, align=left] (mark_des) {Most likely author:};
\node[left=of score1, align=left] (score_des) {Logprob:};
\end{tikzpicture}
\caption{Example of prompt construction and authorship attribution based on log probabilities. The logprob is computed on the orange part, which represents the text from unknown author.}
    \label{fig:example}
\vspace{-10pt}
\end{figure*} 

Our approach to authorship attribution is based on a Bayesian framework. Given a document whose authorship is unknown, our objective is to identify the most probable author from a set using the capabilities of Large Language Models (LLMs).

We consider a scenario where we have a set of authors $\mathcal{A}=\{a_1,\dots,a_n\}$ and a set of all possible texts $\mathcal{E}$. Given an authorship attribution problem, where each author $a_i$ has written a set of texts $t_{i,1},t_{i,2},\dots,t_{i,m_i}\in\mathcal{E}$, we denote the collection of known texts of an author $a_i$ as $\bm{t}(a_i)=(t_{i,1},t_{i,2},\dots,t_{i,m_i})$. For an unknown text $u\in\mathcal{E}$, we aim to determine the most likely author from the set $\mathcal{A}$.

To estimate the author of text $u$, we use a Bayesian framework where the probability that $u$ was written by author $a_i$ is given by:

\begin{equation}
P(a_i|u) = \frac{P(u|a_i)P(a_i)}{P(u)}.
\end{equation}

Here, $P(a_i)$ is the prior probability of each author, assumed to be equal unless stated otherwise, making the problem focus primarily on estimating $P(u|a_i)$.

Assuming that each author $a_i$ has a unique writing style represented by a probability distribution $P(\cdot|a_i)$, texts written by $a_i$ are samples from this distribution. To estimate $P(u|a_i)$, we consider the independence assumption: texts by the same author are independently and identically distributed (i.i.d.). Thus, the unknown text $u$ is also presumed to be drawn from $P(\cdot|a_i)$ for some author $a_i$ and is independent of other texts from that author.

Notice that although texts are independent under the i.i.d. assumption when conditioned on a particular author, there exists a correlation between the unknown text $u$ and the set of known texts $\bm{t}(a)$ in the absence of knowledge about the author. This correlation can be exploited to deduce the most likely author of $u$ using the known texts.

Specifically, we have
\begin{equation}
\begin{aligned}
P(u|\bm{t}(a_i))=&\sum_{a_j\in\mathcal{A}} P(u,a_j|\bm{t}(a_i))
\\=&
\sum_{a_j\in\mathcal{A}} P(u|a_j,\bm{t}(a_i)) P(a_j|\bm{t}(a_i))
\\=&
\sum_{a_j\in\mathcal{A}} P(u|a_j) P(a_j|\bm{t}(a_i)),
\end{aligned}
\end{equation}
where the last equality uses the i.i.d. assumption, meaning that when conditioned on a specific author $a_j$, $u$ is independent of other texts.

We then introduce the "sufficient training set" assumption, where:

\begin{equation}
P(a_j|\bm{t}(a_i))=\begin{cases}1&a_i=a_j\\0&a_i\neq a_j.\end{cases}
\end{equation}

This implies that the training set is sufficiently comprehensive to unambiguously differentiate authors, leading to:

\begin{equation}
P(u|\bm{t}(a_i))=P(u|a_j),
\end{equation}
where $a_j$ is the assumed true author of text $u$.

We use Large Language Models (LLMs) to estimate $P(u|\bm{t}(a_i))$, which represents the probability that a new text $u$ was written by the author of a given set of texts $\bm{t}(a_i)$. 

The probability nature of language models means that they typically calculate the probability of a token or a sequence of tokens given prior context. For a vocabulary set $\Sigma$, the input to a language model might be a sequence of tokens $x_1,\dots,x_m\in\Sigma$, and the model's output would be the probability distribution $P_{\opn{LLM}}(\cdot|x_1,\dots,x_m)$, typically stored in logarithmic scale for numerical stability.

When using an autoregressive language model, we can measure not only the probability of the next token but also the probability of a subsequent sequence of tokens. For instance, if we have a prompt consisting of tokens $x_1,\dots,x_m\in\Sigma$, and we want to measure the probability of a sequence $y_1,\dots,y_s\in\Sigma$, we calculate:

\begin{equation}
\begin{aligned}
&P_{\opn{LLM}}(y_1,\dots,y_s|x_1,\dots,x_m)\\
=&\prod_{i=1}^s P_{\opn{LLM}}(y_i|x_1,\dots,x_m,y_1,\dots,y_{i-1}).
\end{aligned}
\end{equation}

To estimate $P(u|\bm{t}(a_i))$ for authorship attribution, we define:

\begin{equation}
\begin{aligned}
&P(u|\bm{t}(a_i))\\
=&P_{\opn{LLM}}(u|\opn{prompt\_construction}(\bm{t}(a_i))).
\end{aligned}
\end{equation}

The prompt construction can vary, providing flexibility in how we use the model to estimate probabilities. Our method involves constructing a prompt steering the LLM uses to predict the likelihood that the unknown text was written by the same author (\Cref{fig:example}).

In summary, our approach is straightforward and simple. By leveraging the capabilities of Large Language Models, we calculate the likelihood that an unknown text originates from a known author based on existing samples of their writing. This probability assessment allows us to identify the most likely author from a set without the need for fine-tuning or feature engineering.  \section{Experimental Setups}
\subsection{Models \& Baselines} 
\paragraph{Models} We selected two widely-used LLM families: 1) LLaMA family, which includes LLaMA-2~\cite{touvron2023llama}, LLaMA-3, CodeLLaMA~\cite{roziere2023code}, available in various parameter sizes and configurations, with some models specifically fine-tuned for dialogue use cases; 2) the GPT family~\cite{brown2020language}, featuring GPT-3.5-Turbo and GPT-4-Turbo~\cite{achiam2023gpt}, where we specifically used versions gpt-4-turbo-2024-04-09 and gpt-3.5-turbo-0125. The LLaMA family models were deployed using the vLLM framework \cite{kwon2023efficient} if used for Logprob method and are deployed on Azure if used for question-answering. 
Apart from \Cref{tab:main_result}, all ablation studies of Logprob method uses LLaMA-3-70B model.

\paragraph{Baselines} We chose two types of baselines for comparison. 1) embedding-based methods such as BertAA \cite{fabien-etal-2020-bertaa} and GAN-BERT \cite{silva2023authorship}, which require training or fine-tuning, 2) LLM-based methods such as those described in \cite{huang2024can}, which utilize LLMs for authorship attribution tasks through a question-answering (QA) approach.

\subsection{Evaluations} 
\paragraph{Datasets}  We evaluated our method on two widely used author attribution datasets: 1) IMDB62 dataset, a truncated version of IMDB dataset \cite{seroussi2014authorship} and 2) Blog Dataset \cite{schler2006effects}. IMDB62 dataset comprises 62k movie reviews from 62 authors, with each author contributing 1000 samples. Additionally, it also provides some extra information such as the rating score. The Blog dataset, contains 681k blog comments, each with an assigned authorID. Besides the raw text and authorID, each entry includes extra information such as gender and age. Both datasets are accessible via HuggingFace.

\paragraph{Benchmark Construction}
Unlike fixed author sets used in many previous studies, we constructed a random author set for each test to minimize variance. By default, unless specified otherwise, each experiment in our experiments involved a 10-author one-shot setting, and we conducted 100 tests for each experiment to reduce variance. Each test involved the following steps: 1) Ten candidate authors were randomly selected. 2) For each author, one (or n for n-shot) article was randomly selected as the training set. 3) One author was randomly selected from the ten candidates as the test author. 4) One article not in the training set was randomly selected from the test author’s articles as the test set (with size of 1). 5) We run the authorship attribution algorithm to classify the test article into 10 categories.

Our evaluation pipeline can avoid potential biases from fixed author sets and better measure the efficacy of LLMs in authorship attribution tasks. We also share our pipeline for fair evaluations of future related works. 

Notably, aforementioned pipeline is suitable for non-training based methods like ours and QA approaches. However, for training-based methods such as embedding approaches, each train-test split is followed by a retraining, demanding significant computational resources. Therefore, in this work, we directly cited scores from the original papers.

\paragraph{Evaluation Metrics} We adopt three metrics: top-1, top-2 and top-5 accuracies. Specifically, top k accuracy is computed as follows:
\myvspace{-4pt}
\begin{equation}
    \mathrm{Top} \, k \, \mathrm{Accuracy} = \frac{\mathrm{\mathrm{Num}_{\mathrm{correct}}^k}}{\mathrm{Num}_{\mathrm{all}}},
\myvspace{-4pt}
\end{equation}
where $\mathrm{\mathrm{Num}_{\mathrm{correct}}^k}$ represents the number of tests where the actual author is among the top k predictions, and $\mathrm{Num}_{\mathrm{all}}$ represents the total number of tests. \section{Experiments}
Firstly, we evaluate different methods for author attribution in \Cref{se:performance}, noting that our Logprob method significantly outperformed QA-based methods in accuracy and stability across datasets. Then, we study the impact of increasing candidate numbers on performance in \Cref{se:num_candidate}, where our method maintained high accuracy despite a larger pool of candidates. Next, in \Cref{se:prompt_sensitivity}, we analyze prompt sensitivity, concluding that while prompt use is crucial, variations in prompt design did not significantly affect the performance. Further, in \Cref{se:bias}, we explore bias in author attribution and in \Cref{se:subgroup}, we measure performance variations across different subgroups. Finally, in \Cref{se:efficiency}, we compared the efficiency of different author attribution methods.

\subsection{Author Attribution Performance}\label{se:performance}
\begin{table*}[t!]
\small
\setlength{\tabcolsep}{2.5pt}
\centering
\resizebox{\textwidth}{!}{\begin{tabular}{llcccccccccc}
\toprule
\multirow{2}{*}{\textbf{Method}}  & \multirow{2}{*}{\textbf{Model}}  & \multicolumn{5}{c}{\textbf{IMDB62 Dataset}} &\multicolumn{5}{c}{\textbf{BLOG Dataset}} \\ 
\cmidrule(r){3-7} \cmidrule(r){8-12}
 & &\textbf{\#Candidate} & \textbf{n-Shot} & \textbf{Top 1 Acc.} & \textbf{Top 2 Acc.} & \textbf{Top 5 Acc.} &\textbf{\#Candidate} & \textbf{n-Shot} & \textbf{Top 1 Acc.} & \textbf{Top 2 Acc.} & \textbf{Top 5 Acc.}  \\ 
\midrule
\multirow{12}{*}{LogProb} & LLaMA-2-7B & 10& 1 & 80.0 $\pm$ 4.0 & 88.0 $\pm$ 3.3 & 97.0 $\pm$ 1.7 & 10& 1 & 79.0 $\pm$ 4.1 & 84.0 $\pm$ 3.7 & \textbf{98.0 $\pm$ 1.4} \\
& LLaMA-2-7B-Chat & 10 & 1 & 68.0 $\pm$ 4.7 & 80.0 $\pm$ 4.0 & 88.0 $\pm$ 3.3 & 10 & 1 & 69.0 $\pm$ 4.6 & 78.0 $\pm$ 4.1 & 89.0 $\pm$ 3.1 \\
& LLaMA-2-13B  & 10 & 1 & 84.0 $\pm$ 3.7 & 88.0 $\pm$ 3.3 & \textbf{100.0 $\pm$ 0.0} & 10 & 1 & 81.0 $\pm$ 3.9 & 86.0 $\pm$ 3.5 & 94.0 $\pm$ 2.4 \\
& LLaMA-2-70B  & 10 & 1 & \textbf{88.0 $\pm$ 3.3} & \textbf{94.0 $\pm$ 2.4} & \textbf{99.0 $\pm$ 1.0} & 10 & 1 & \textbf{88.0 $\pm$ 3.3} & \textbf{90.0 $\pm$ 3.0} & 95.0 $\pm$ 2.2 \\ 
 & LLaMA-2-70B-Chat & 10 & 1 & 79.0 $\pm$ 4.1 & 85.0 $\pm$ 3.6 & 95.0 $\pm$ 2.2 & 10 & 1 & 83.0 $\pm$ 3.8 & 85.0 $\pm$ 3.6 & \textbf{97.0 $\pm$ 1.7}  \\ 
& Code-LLaMA-7B & 10 & 1 & 71.0 $\pm$ 4.5 & 84.0 $\pm$ 3.7 & 96.0 $\pm$ 2.0 & 10 & 1 & 78.0 $\pm$ 4.1 & 84.0 $\pm$ 3.7 & 94.0 $\pm$ 2.4 \\
& Code-LLaMA-13B & 10 & 1 & 70.0 $\pm$ 4.6 & 84.0 $\pm$ 3.7 & 98.0 $\pm$ 1.4 & 10 & 1 & 77.0 $\pm$ 4.2 & 85.0 $\pm$ 3.6 & 92.0 $\pm$ 2.7 \\
& Code-LLaMA-34B & 10 & 1 & 75.0 $\pm$ 4.3 & 84.0 $\pm$ 3.7 & 98.0 $\pm$ 1.4 & 10 & 1 & 78.0 $\pm$ 4.1 & 83.0 $\pm$ 3.8 & 94.0 $\pm$ 2.4 \\
& LLaMA-3-8B  & 10 & 1 & 82.0 $\pm$ 3.8 & 89.0 $\pm$ 3.1 & 98.0 $\pm$ 1.4 & 10 & 1 & 84.0 $\pm$ 3.7 & \textbf{89.0 $\pm$ 3.1} & 95.0 $\pm$ 2.2 \\
& LLaMA-3-8B-Instruct & 10 & 1 & 69.0 $\pm$ 4.6 & 77.0 $\pm$ 4.2 & 90.0 $\pm$ 3.0 & 10 & 1 & 68.0 $\pm$ 4.7 & 77.0 $\pm$ 4.2 & 90.0 $\pm$ 3.0 \\
& LLaMA-3-70B & 10 & 1 & \textbf{85.0 $\pm$ 3.6} & \textbf{93.0 $\pm$ 2.6} & 98.0 $\pm$ 1.4 & 10 & 1& 82.0 $\pm$ 3.8 & \textbf{88.0 $\pm$ 3.3} & 95.0 $\pm$ 2.2\\ 
& LLaMA-3-70B-Instruct & 10 & 1 & 79.0 $\pm$ 4.1 & 89.0 $\pm$ 3.1 & \textbf{99.0 $\pm$ 1.0} & 10 & 1 & 79.0 $\pm$ 4.1 & \textbf{87.0 $\pm$ 3.4} & \textbf{96.0 $\pm$ 2.0} \\ 
\midrule
\multirow{4}{*}{\makecell[l]{QA}} & LLaMA-2-70B-Chat & 10 & 1 & Failed & - & - & 10 & 1 &Failed & - & - \\
& LLaMA-3-70B-Instruct & 10 & 1 & 31.0 $\pm$ 4.6 & - & - & 10 & 1 & 22.0 $\pm$ 4.1 & - & - \\
& GPT-3.5-Turbo  & 10 & 1 & 69.0 $\pm$ 4.6 & - & - & 10 & 1 & 47.0 $\pm$ 5.0  & - & - \\ 
 & GPT-4-Turbo  & 10 & 1 & 34.0 $\pm$ 4.7 & - & - & 10 & 1 & 62.0 $\pm$ 4.9 & - & - \\
 \midrule
 \multirow{2}{*}{Other Baseline} &  GAN-BERT   & 20 & 80 &  96.0 & - & - & 20 & 80 & 40.0 & - & - \\ 
& BertAA & 62 & 80 &  93.0 & - & - & 10 & 80 & 65.0 & - & - \\
\bottomrule
\end{tabular}}
\caption{Author attribution results on IMDB62 and Blog dataset. Prompt construction for QA method is in consistent with \citet{huang2024can}. }
\label{tab:main_result}
\end{table*} \Cref{tab:main_result} shows the main results for different methods on the IMDB62 and Blog datasets concerning authorship attribution capabilities. We make the following observations: 

\begin{itemize}[leftmargin=0.15in]
\item \textbf{LLMs with QA-based methods cannot perform author attribution tasks effectively.}
For example, GPT-4-Turbo can only achieve a top-1 accuracy of 34\% on the IMDB62 dataset and 62\% on the Blog dataset. Notably, there are two interesting phenomena: 1) GPT-4-Turbo and GPT-3.5-Turbo exhibit inconsistent higher accuracy across different datasets, highlighting inherent instability in the prompt-based approach. 2) Older LLMs with smaller context window lengths are unable to perform author attribution due to the prompt exceeding the context window. These phenomena indicate that QA methods are not a good option for enabling LLMs to conduct author attribution tasks effectively.

\item \textbf{Our Logprob method helps LLMs perform author attribution tasks more effectively.}
With LLaMA-3-70B, we achieved top-1 accuracy of 85\%, and both top-2 and top-5 accuracies were even higher. This suggests that LLMs equipped with our method can effectively narrow down large candidate sets. Additionally, two another things worth noting are that 1) LLMs with the Logprob method exhibit more stable performance across both tasks, something QA methods struggle with, and 2) LLMs with Logprob can conduct authorship attribution tasks with lower requirements for context window length. For instance, LLaMA-2-70B-Chat with the Logprob method can handle authorship attribution, whereas the same model with a QA approach fails when the collective text of 10 authors exceeds the context window length. These findings highlight the superiority of our Logprob method.

\item \textbf{Training-free method can achieve comparable or even superior performance to training-based methods}. The Blog dataset showed higher top-1 accuracy with LLaMA + Logprob compared to GAN-BERT and BertAA. While the IMDB62 dataset exhibited lower performance relative to embedding-based methods, it is important to note that Logprob achieves this as a one-shot method, whereas embedding-based approaches require much more data for training to converge. This demonstrates that Logprob can more effectively capture the nuances necessary for authorship attribution.
\end{itemize}

\subsection{Performance vs. Number of Candidates}\label{se:num_candidate}
\begin{figure}[t!]
    \centering
    \tikzset{global scale/.style={
        scale=#1,
        every node/.append style={scale=#1}
}}
    \begin{tikzpicture}[]
        \pgfplotsset{set layers}
        \scriptsize{
            \begin{axis}[
                align=center,
                at={(0,0)},
                ymajorgrids,
                xmajorgrids,
                axis lines=left,
                axis x line*=bottom,
                axis y line*=left,
                grid style=dashed,
                width=0.45\textwidth,
                height=.3\textwidth,
                xlabel={\small \# Candidates},
                ylabel={\small{Accuracy (\%)}},
                ylabel style={yshift=-2em},
                xlabel style={yshift=1.0em},
                yticklabel style={/pgf/number format/precision=0,/pgf/number format/fixed zerofill},
                ymin=60,
                ymax=105,
                ytick={70, 80, 90, 100},
                xmin=-3,
                xmax=65,
                xtick={2, 5, 10, 20, 50},
                legend style={
                    at={(0,0)},
                    anchor=north east,
                    at={(axis description cs:0.3,0.42)},
                    fill=none,
                    draw=none,
                    yshift=-0.5em,
                    xshift=0.5em,
                    inner sep=0pt,
                    legend plot pos=left,
                    font={\small},
                    cells={anchor=west},
                    legend columns=1,
                    column sep=0pt
                }
            ]
                \addplot+[blue, mark=*, mark size=1.0pt, thick, mark options={solid, fill=white, draw=blue, line width=1.pt}, error bars/.cd,
                    y dir=plus, y explicit, y dir=both, error bar style={mark size=2.5pt, thick, color=blue}]
                    coordinates {
                        (2, 100) +- (0,0)
                        (5, 100) +- (0,0.0) 
                        (10, 98) +- (0,1.4)
                        (20, 93) +- (0,2.6)
                        (50, 87) +- (0,3.4)
                    };\addlegendentry{\small{Top 5}}

                \addplot+[red, mark=star, mark size=2.5pt, thick, mark options={solid, fill=red, draw=red, line width=1.pt}, error bars/.cd,
                    y dir=plus, y explicit, y dir=both, error bar style={thick, mark size=2.5pt, color=red}]
                    coordinates {
                        (2, 100) +- (0,0)
                        (5, 96) +- (0,2.0) 
                        (10, 93) +- (0,2.6)
                        (20, 87) +- (0,3.4)
                        (50, 84) +- (0,3.7)
                    };\addlegendentry{\small{Top 2}}

                \addplot+[purple, mark=*, mark size=1pt, thick,
                    mark options={solid, fill=purple, draw=purple, line width=1.pt}, error bars/.cd, 
                    y dir=plus, y explicit, y dir=both, error bar style={mark size=2.5pt, thick, color=purple}]
                    coordinates {
                        (2, 97) +- (0,1.7)
                        (5, 91) +- (0,2.9) 
                        (10, 85) +- (0,3.6)
                        (20, 81) +- (0,3.9)
                        (50, 76) +- (0,4.3)
                    };\addlegendentry{\small{Top 1}}
            \end{axis}
        }
    \end{tikzpicture}
    \caption{Accuracy vs. number of candidates.}
    \label{fig:performance_vs_candidates}
\end{figure}
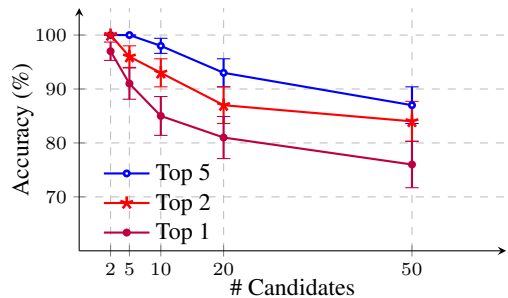
 \begin{table*}[t!]
\small
\centering
\setlength{\tabcolsep}{5pt}
\begin{tikzpicture}
    \definecolor{ugreen}{RGB}{118,218,145}
    \node[align=right,left] (Prompt2) at (0,0) {\makecell[l]{\textbf{Prompt 1}: Here is the text from the same author: \\ \textbf{Prompt 2}: \textcolor{red}{Analyze the writing styles of the input texts, disregarding the differences in topic and content.} \\ Here is the text from the same author: \\ \textbf{Prompt 3}: \textcolor{blue}{Focus on grammatical styles indicative of authorship.} Here is the text from the same author: \\ \textbf{Prompt 4}: \textcolor{ugreen}{Analyze the writing styles of the input texts, disregarding the differences in topic and content.} \\ \textcolor{ugreen}{Reasoning based on linguistic features such as phrasal verbs, modal verbs, punctuation, rare words, affixes,} \\ \textcolor{ugreen}{quantities, humor, sarcasm, typographical errors, and misspellings.} Here is the text from the same author:}};
\end{tikzpicture}

\begin{tabular}{lcccc}
\toprule
\# &\bf Prompting & \bf Top 1 Accuracy & \bf Top 2 Accuracy & \bf Top 5 Accuracy \\
\midrule
1 &<Example Text> + <Query Text> & 70.0 $\pm$ 4.6 & 81.0 $\pm$ 3.9 & 92.0 $\pm$ 2.7 \\
2 &<Example Text> + <Prompt 1> + <Query Text> & 85.0 $\pm$ 3.6 & 92.0 $\pm$ 2.7 & 99.0 $\pm$ 1.0 \\
3 &<Example Text> + <Prompt 2> + <Query Text> & 83.0 $\pm$ 3.8 & 87.0 $\pm$ 3.4 & 100.0 $\pm$ 0.0 \\
4 &<Example Text> + <Prompt 3> + <Query Text> & 86.0 $\pm$ 3.5 & 90.0 $\pm$ 3.0 & 100.0 $\pm$ 0.0 \\
5 &<Example Text> + <Prompt 4> + <Query Text> & 87.0 $\pm$ 3.4 & 90.0 $\pm$ 3.0 & 99.0 $\pm$ 1.0
 \\
\bottomrule
\end{tabular}
\caption{Author attribution performance vs. different prompting choices on IMDB62 dataset.}
\label{tab:performance_vs_prompt}
\end{table*} \begin{table}[t!]
\small
\centering
\setlength{\tabcolsep}{1.5pt}
\begin{tabular}{cccc}
\toprule
\bf Gender & \bf Top 1 Acc. & \bf Top 2 Acc. & \bf Top 5 Acc. \\
\midrule
 Both & 84.0 $\pm$ 1.6 & 90.8 $\pm$ 1.3 & 95.8 $\pm$ 1.0\\
 \midrule
 Male & 81.4 $\pm$ 2.5 & 88.6 $\pm$ 2.1 & 95.4 $\pm$ 1.4 \\
 Female & \bf 86.3 $\pm$ 2.1 & \bf 92.8 $\pm$ 1.6 & \bf 96.2 $\pm$ 1.2\\
\bottomrule
\end{tabular}
\caption{Gender bias in author attribution performance.}
\label{tab:bias_performance_vs_gender}
\end{table} One of the challenges in authorship attribution is the difficulty in correctly identifying the author as the number of candidates increases, which generally leads to decreased accuracy. \Cref{fig:performance_vs_candidates} shows the author attribution performance across different candidate counts on the IMDB62 dataset. We made the following observations: 

\begin{itemize}[leftmargin=0.15in]
\item First, performance indeed decreases as the number of candidates increases.
\item Second, across all settings, all metrics maintain relatively high scores. For example, in the setting with 50 candidates, our method achieved 76\% top-1 accuracy, 84\% top-2 accuracy, and 87\% top-5 accuracy.
\item Third, top-2 and top-5 accuracies are more stable compared to top-1 accuracy. The model may not always place the correct author at the top, but it often includes the correct author within the top few predictions. This attribute is also crucial as it allows the narrowing down of a large pool of candidates to a smaller subset of likely candidates.
\end{itemize}

\subsection{Analysis of Prompt Sensitivity}\label{se:prompt_sensitivity}
Our method relies on suitable prompt as in \Cref{fig:example}. Here, we discuss the sensitivity of our accuracy to different prompt constructions in \Cref{tab:performance_vs_prompt}. We made the following observations:

\begin{itemize}[leftmargin=0.15in]
\item \textbf{Using prompts is essential for enhancing the accuracy of our method (\#1 vs. \#2).} This phenomenon is aligned with previous studies \cite{wei2022chain} that have demonstrated that prompting is beneficial for unlocking the full potential of LLMs.

\item \textbf{There is no statistically significant evidence to suggest that specific prompt designs impact performance significantly (\#2 vs. \#3 vs. \#4 vs. \#5).} The results show very close performance metrics across different prompt constructions.
\end{itemize}

\paragraph{Discussions} Prompting sensitivity \cite{sclar2023quantifying} is a widely acknowledged property in the generation process of LLMs. This also has motivated a trend of research on prompting engineering \cite{zhang2023automatic, guo2024connecting} as different promptings can lead to completely different performance. However, our method appears to be relatively insensitive to the choice of prompt, which makes our method more robust, maintaining high performance and stability across various settings.

\subsection{Bias Analysis}\label{se:bias}
\begin{table}[t!]
\small
\centering
\setlength{\tabcolsep}{1.5pt}
\begin{tabular}{cccc}
\toprule
\bf Gender & \bf Top 1 Acc. & \bf Top 2 Acc. & \bf Top 5 Acc. \\
\midrule
 Male & 77.0 $\pm$ 4.2 & 82.0 $\pm$ 3.8 & 92.0 $\pm$ 2.7 \\
 Female & \bf 89.0 $\pm$ 3.1 & \bf 91.0 $\pm$ 2.9 & \bf 95.0 $\pm$ 2.2\\
\bottomrule
\end{tabular}
\caption{Author attribution performance in each gender subgroup.}
\label{tab:subgroup_gender}
\end{table} An algorithm trained on an entire dataset may exhibit different accuracy levels across different subgroups during testing \citep{chouldechova2017fairer,pastor2021identifying}. This section discusses such bias issues and measures how the algorithm's accuracy varies for different subgroups. 
\begin{table*}[t!]
    \centering
\small
\begin{minipage}[b]{0.47\textwidth}
\centering
\setlength{\tabcolsep}{1.5pt}
\renewcommand{\arraystretch}{1}
\resizebox{0.7\linewidth}{!}{
\begin{tabular}{cccc}
\toprule
\bf Interval & \bf Top 1 Acc. & \bf Top 2 Acc. & \bf Top 5 Acc. \\
\midrule
 $\left[1-2\right]$ & 82.0 $\pm$ 3.8 & 89.0 $\pm$ 3.1 & 96.0 $\pm$ 2.0 \\
 $\left[3-4\right]$ &  87.0 $\pm$ 3.4 & 94.0 $\pm$ 2.4 & 99.0 $\pm$ 1.0 \\
 $\left[5-6\right]$ & \bf 90.0 $\pm$ 3.0 & \bf 96.0 $\pm$ 2.0 & \bf 100.0 $\pm$ 0.0 \\
 $\left[7-8\right]$ & 88.0 $\pm$ 3.3 & 92.0 $\pm$ 2.7 & 97.0 $\pm$ 1.7  \\
 $\left[9-10\right]$ & 89.0 $\pm$ 3.1 & 93.0 $\pm$ 2.6 & 96.0 $\pm$ 2.0\\
\bottomrule
\end{tabular}
}
\caption*{(a) performance in each rating subgroup.}
\label{tab:subgroup_rating}
\end{minipage}
\hspace{-0.5in}
\begin{minipage}[b]{0.47\textwidth}
\centering
\setlength{\tabcolsep}{1.5pt}
\renewcommand{\arraystretch}{1.2}
\resizebox{0.7\linewidth}{!}{
\begin{tabular}{cccc}
\toprule
\bf Age & \bf Top 1 Acc. & \bf Top 2 Acc. & \bf Top 5 Acc. \\
\midrule
 $\left[13-17\right]$ & \bf 90.0 $\pm$ 3.0 & \bf 94.0 $\pm$ 2.4 & \bf 99.0 $\pm$ 1.0 \\
 $\left[18-34\right]$ & 84.0 $\pm$ 3.7 & 89.0 $\pm$ 3.1 & 95.0 $\pm$ 2.2 \\
 $\left[35-44\right]$ & 80.0 $\pm$ 4.0 & 87.0 $\pm$ 3.4 & 94.0 $\pm$ 2.4\\
 $\left[45-48\right]$ & 81.0 $\pm$ 3.9 & 85.0 $\pm$ 3.6 & 95.0 $\pm$ 2.2 \\
\bottomrule
\end{tabular}}
\caption*{(b) performance in each age subgroup.}
\label{tab:subgroup_age}
\end{minipage}
\myvspace{-5pt}
\caption{Author attribution performance in each rating subgroup and age subgroup.}
\label{tab:subgroup_rating_age}
\myvspace{-5pt}
\end{table*} \begin{table*}[ht]
\small
\centering
\setlength{\tabcolsep}{3pt}
\begin{tabular}{cccccc}
\toprule
\# &\bf Foundation Models & \bf Deployment Resource & \bf Method & \bf Inference Time (s) & \bf Accuracy \\
\midrule
1 & LLama-3-70B & 8 $\times$ A6000 (VLLM) & Logprob & 462.1 & 85.0 $\pm$ 3.6\\
2 &GPT-4-Turbo& OpenAI& QA & 663.1 &  34.0 $\pm$ 4.7 \\
3 & LLama-3-70B-Instruct & Azure & QA & 2065.6 & 31.0 $\pm$ 4.6\\
\bottomrule
\end{tabular}
\myvspace{-5pt}
\caption{Efficiency analysis between prompt-based method and logprob-based method on Blog dataset.}
\label{tab:efficient_analysis}
\myvspace{-5pt}
\end{table*} \paragraph{Influence of Gender}\label{sec:bias_gender}
We conduct 500 tests which consists of 237 tests for blogs written by male authors and 263 tests for blogs written by female authors and show their accuracy of authorship attribution separately in 
\Cref{tab:bias_performance_vs_gender}. The results indicate that authorship attribution for blogs written by female authors exhibits higher accuracy. This suggests that female-authored blogs might contain more distinct personal styles, making it easier to infer the author.

\subsection{Subgroup Analysis}\label{se:subgroup}
When considering authorship attribution restricted to specific subgroups, the task can either become simpler or more difficult. Certain subgroups may express personal styles more distinctly, making authorship attribution easier, while others may be more homogeneous, making it more challenging. Here, we consider three subgroup factors: gender, age, and rating, to analyze the performance under each group.

\paragraph{Subgroup by Gender}
As shown in \Cref{tab:subgroup_gender}, we evaluated the performance of authorship attribution within different gender subgroups in the Blog dataset. We observed that authorship attribution performed better within the female subgroup, consistent with findings in \Cref{sec:bias_gender}, suggesting female-authored blogs possess more distinctive personal styles.

\paragraph{Subgroup by Rating}
\Cref{tab:subgroup_rating_age} (a) shows the performance of authorship attribution across different rating ranges in the IMDb review dataset. Overall, we can see that rating does influence performance, with review in the $\left[5-6\right]$ rating range easier to attribute. Despite such difference, our method consistently obtains good performance across all subgroups.

\paragraph{Subgroup of Age}
\Cref{tab:subgroup_rating_age} (b) shows the performance of authorship attribution across different age ranges of bloggers in the Blog dataset. We observed that age significantly influences performance. The youngest age group $\left[13-17\right]$ exhibited the highest top-1 accuracy at 90\%, while accuracy decreased with increasing author age. This suggests that younger authors tend to have more distinct opinions and identifiable writing styles. Despite performance differences, our method maintained relatively overall high performance, with the lowest accuracy still surpassing that of GPT-4-Turbo with QA method.

\subsection{Efficiency Analysis}\label{se:efficiency}
\Cref{tab:efficient_analysis} shows the efficiency comparison of different methods on the imdb dataset. Our Logprob method operates with notably lower runtime compared to QA methods. This is primarily due to the Logprob method requiring only a single forward pass through the LLM for each author to estimate the log probabilities. In contrast, QA methods generally need multiple iterations of token generations to form a response, which increases computation time substantially. In the mean time, our method achieves an accuracy of up to 85\%, surpassing QA method based on GPT-4-Turbo in both efficiency and accuracy.

In summary, our method proves to be effective and efficient in performing authorship attribution across various datasets and setups. \section{Conclusion}
In this paper, we study the problem of authorship attribution. We demonstrate the effectiveness of utilizing pre-trained Large Language Models (LLMs) for one-shot author attribution. Our Bayesian approach leverages the probabilistic nature of language models like Llama-3 to infer authorship. Our method does not require fine-tuning, therefore reduces computational overhead and data requirements. Our experiments validate that our method is more effective and efficient compared to existing techniques.  \section{Limitations}
The main limitations arise due to the dependence on LLMs. 

Our method relies heavily on the capabilities of LLMs, and the performance of our approach is highly affected by the size and training objectives of the LLMs. As shown in \Cref{tab:main_result}, models that are only pre-trained rather than fine-tuned for dialogue or code task performs better.

While larger models generally perform better, they also entail higher costs, posing scalability and accessibility challenges for broader applications.

Another limitation is due to training data of LLMs. If the training data lacks diversity or fails to include certain writing styles, the model may not fully capture the intricacies of an author's style, potentially leading to misclassifications. This limitation underscores the importance of using diverse and comprehensive training datasets.

Furthermore, any biases present in the training data can also be absorbed by the model. These biases will influence the performance of our authorship attribution method.

On the broader societal level, the potential for misuse of this technology is a significant concern. The challenge of regulating and overseeing the use of such powerful tools is still not fully addressed.

Lastly, while our approach avoids the need for extensive retraining or fine-tuning, which is an advantage in many cases, this also means that our method might not adapt well to scenarios where lots of training data and computation is available, which justifies more complex and computationally intensive methods. \section*{Acknowledgments}
ZH, TZ and HH were partially supported by NSF IIS 2347592, 2347604, 2348159, 2348169, DBI 2405416, CCF 2348306, CNS 2347617.

\bibliography{content/references}

\appendix

\section{Ethical Considerations}

Our method using LLMs for authorship attribution brings several ethical considerations that must be addressed to ensure responsible and fair use of the technology.

\paragraph{Privacy and Anonymity}
The capacity of LLMs to attribute authorship with high accuracy can lead to ethical challenges regarding privacy and anonymity. Individuals who wish to remain anonymous or protect their identity could be compromised if authorship attribution tools are misused. Therefore, it is crucial to establish strict guidelines and ethical standards on the use of such technologies to prevent breaches of privacy.

\paragraph{Potential for Abuse}
Despite multiple beneficial applications, the misuse potential of authorship attribution tools is significant. Risks include the use of this technology to suppress free speech or to endanger personal safety by identifying individuals in contexts where anonymity is crucial for safety. Addressing these risks requires robust governance to prevent misuse and to ensure that the technology is used ethically and responsibly.

\paragraph{Bias Issue}
The performance of authorship attribution methods can vary across different demographics, leading to potential biases. It is important to continually assess and correct these biases to ensure fairness in the application of this technology. 

\paragraph{Misclassification Issue}
Given the high stakes involved, especially in forensic contexts, the accuracy of authorship attribution is important. Misclassifications can have serious consequences, including wrongful accusations or legal implications. It is essential for authorship attribution methods to be reliable and for their limitations to be transparently communicated to users.

\section{Broader Impact}

Our study of authorship attribution using LLMs contributes to advancements in various domains:

\paragraph{Forensic Linguistics} Our research contributes to the field of forensic linguistics by providing tools that can solve crimes involving anonymous or disputed texts. This can be particularly useful for law enforcement and legal professionals who need to gather evidence and make more informed decisions.

\paragraph{Intellectual Property Protection} Our method can serve as a powerful tool in identifying the authors of texts, which can help protect intellectual property rights and resolve disputes in copyright.

\paragraph{Historical Text Attribution} In literary and historical studies, determining the authorship of texts can provide insights into their origins and contexts, enhancing our understanding and interpretation.

\paragraph{Enhanced Content Management} Media and content companies can use this technology to manage content more effectively by accurately attributing authorship to various contributors. 

\paragraph{Educational Applications} In educational settings, our method can help prevent plagiarism and promote academic integrity. It can also serve as a teaching tool to help students understand and appreciate stylistic differences between authors.

While our method holds promise across multiple applications, it is crucial to deploy it with caution. Ensuring that the technology is used responsibly and ethically will be key to maximizing its benefits while minimizing potential harm. 
\end{document}